\documentclass{article}

\usepackage[numbers]{natbib}
\usepackage[preprint]{neurips_2023}

\usepackage{xcolor}         
\definecolor{linkColor}{rgb}{0.18,0.39,0.62}
\usepackage[utf8]{inputenc} 
\usepackage[T1]{fontenc}    
\usepackage[colorlinks=true,linkcolor=linkColor,citecolor=linkColor,filecolor=linkColor,urlcolor=linkColor]{hyperref}       
\usepackage{url}            
\usepackage{booktabs}       
\usepackage{amsfonts}       
\usepackage{nicefrac}       
\usepackage{microtype}      
\usepackage[most]{tcolorbox}
\usepackage{enumitem}
\usepackage{bbm}

\usepackage{amsmath}
\usepackage{amssymb}
\usepackage{mathtools}
\usepackage{amsthm}
\usepackage{graphicx}
\usepackage{subfigure}
\usepackage{xspace}
\usepackage{pifont}

\theoremstyle{plain}

\theoremstyle{definition}

\theoremstyle{remark}

\usepackage[capitalize,noabbrev]{cleveref}
\usepackage{framed}
\usepackage{multirow}
\usepackage{svg}
\usepackage{listings}
\usepackage{wrapfig}
\usepackage{fdsymbol}

\newcommand\ours{\textsc{LongNet}}
\newcommand\our{\textsc{LongNet}}

\title{\ours{}: Scaling Transformers to \\ 1,000,000,000 Tokens}

\author{
\vspace{-0.25in} \\
\textbf{
Jiayu Ding$^{\clubsuit}$\thanks{~Equal contribution. $\dagger$ Corresponding author.}~~~~Shuming Ma$^{\spadesuit}$\footnotemark[1]~~~~Li Dong$^{\spadesuit}$~~~Xingxing Zhang$^{\spadesuit}$} \\
\textbf{
Shaohan Huang$^{\spadesuit}$~~~Wenhui Wang$^{\spadesuit}$~~~{Nanning Zheng}$^{\clubsuit}$$^{\dagger}$~~~{Furu Wei}$^{\spadesuit}$$^{\dagger}$} \\
$^{\spadesuit}$Microsoft Research \\
$^{\clubsuit}$Xi'an Jiaotong University \\
{\href{https://aka.ms/GeneralAI}{https://aka.ms/GeneralAI}}
\vspace{-0.4cm}
\\}

\begin{document}

\maketitle

\begin{abstract}
Scaling sequence length has become a critical demand in the era of large language models.
However, existing methods struggle with either computational complexity or model expressivity, rendering the maximum sequence length restricted.
To address this issue, we introduce \ours{}, a Transformer variant that can scale sequence length to more than 1 billion tokens, without sacrificing the performance on shorter sequences.
Specifically, we propose dilated attention, which expands the attentive field exponentially as the distance grows.
\ours{} has significant advantages: 1) it has a linear computation complexity and a logarithm dependency between any two tokens in a sequence;
2) it can be served as a distributed trainer for extremely long sequences;
3) its dilated attention is a drop-in replacement for standard attention, which can be seamlessly integrated with the existing Transformer-based optimization.
Experiments results demonstrate that \ours{} yields strong performance on both long-sequence modeling and general language tasks.
Our work opens up new possibilities for modeling very long sequences, e.g., treating a whole corpus or even the entire Internet as a sequence.
Code is available at \url{https://aka.ms/LongNet}.
\end{abstract}

\hfill

\begin{figure}[h]
\centering
\includegraphics[width=1.0\columnwidth]{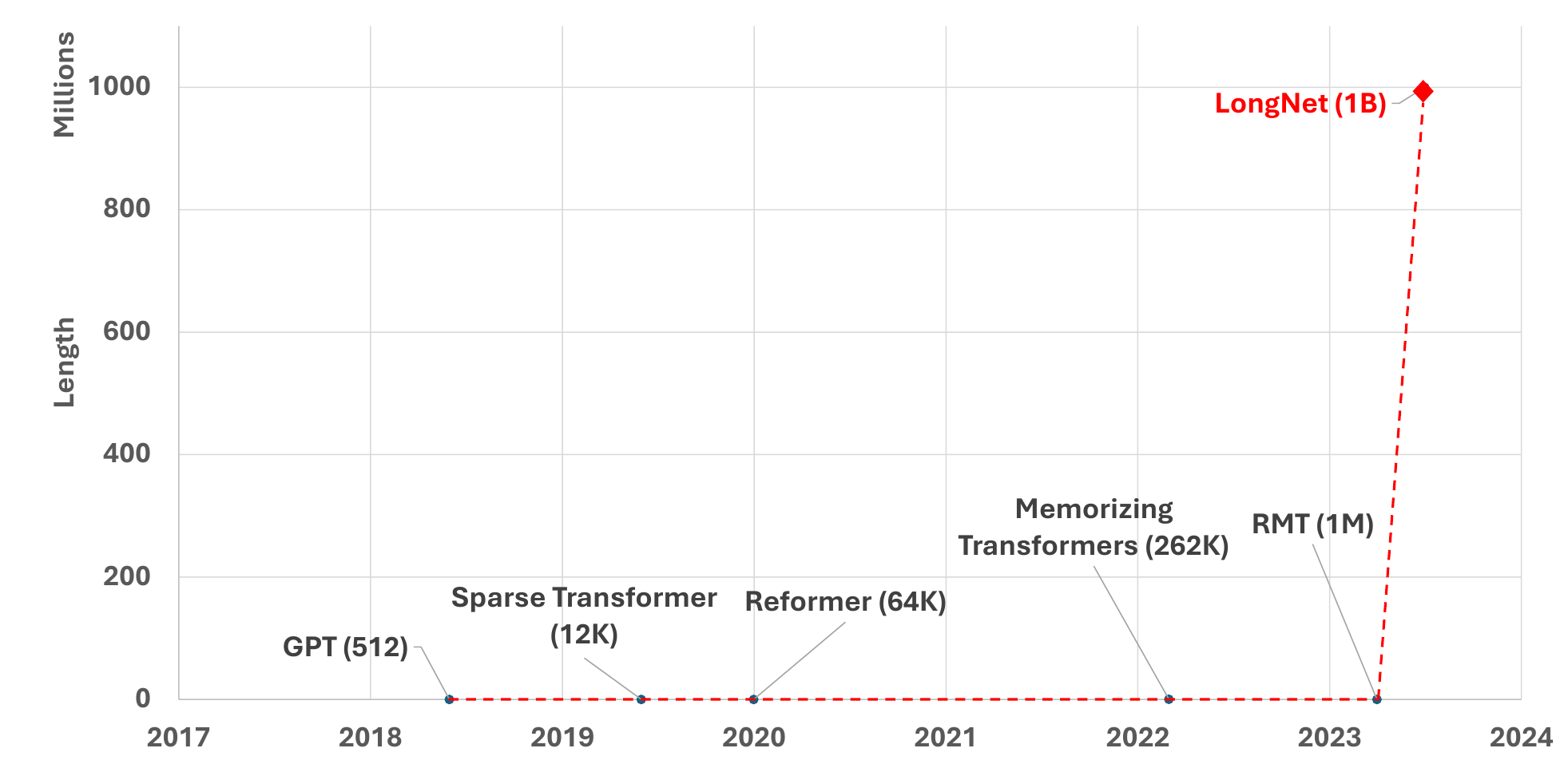}
\caption{Trend of Transformer sequence lengths over time.
}
\vspace{-0.5cm}
\label{fig:trend}
\end{figure}

\newpage

\section{Introduction}
\label{sec:intro}

Recent years have witnessed a trend toward scaling neural networks~\cite{gpt3,scalinglaw,scalingvit,palm,vit22b}. The depth is primarily scaled up for exponential expressivity, producing many powerful deep networks~\cite{resnet,gpipe,deepnet}. Then, the sparse MoE models~\cite{gshard,switch,stmoe} and model parallelism approaches~\cite{megatron,reduceact} efficiently enlarge the hidden dimension.
\textbf{Sequence length, as the last atomic dimension of the neural network, is desirable to be unlimited}.
Breaking the limitation of sequence length introduces significant advantages. First, it provides large memory and receptive field for models, which is practical for them to interact with human and the world. Second, a longer context contains more complex causality and reasoning paths that models can exploit in training data. In contrast, short dependency has more spurious correlations, which is harmful to generalization. Third, it enables to explore the limits of in-context learning, which has the potential to be a paradigm shift for many-shot learning, as an extremely long context may help the models alleviate catastrophic forgetting.

The major challenge of scaling up sequence length is striking the right balance between the computational complexity and the model expressivity. RNN-style models are primarily implemented to increase the length. However, its sequential nature limits the parallelization during training, which is essential in long-sequence modeling. More recently, state space models~\cite{s4,s5,h3,hyena} are appealing to sequence modeling. It can operate as a CNN during training, and transform to an efficient RNN at test time. While they perform well at long-range benchmarks~\cite{lra}, their performance on regular lengths is not as good as Transformers, limited mainly by the model expressivity~\cite{blockstate}. 

Another strand of scaling the sequence length is to decrease the complexity of Transformers, i.e., the quadratic complexity of self-attention. Implementing sliding windows or convolution modules over the attention is a straightforward way to make the complexity nearly linear. Nevertheless, this sacrifices the ability to recall the early tokens, forgetting the prompts at the very beginning of the sequence. Sparse attention reduces the computation by sparsifying the attention matrix, preserving the possibility of recalling long-distant information. For example, \cite{sparsetransformer} obtains $\mathcal{O}(N\sqrt{N}d)$ time complexity with a fixed sparse pattern. Besides the heuristic patterns~\cite{bigbird,longformer}, the learnable patterns prove to be useful for sparse attention~\cite{reformer,colt5}. There are also some other efficient Transformer-based variants, including low-rank attention~\cite{linformer,low-rank}, kernel-based methods~\cite{lineartransformer,performer,normformer}, downsampling approaches~\cite{settransformer,perceiver,luna}, recurrent models~\citep{trm-xl,rmt}, and retrieval-based methods~\citep{memorizingtrm,longmem}. Yet, none has been scaled to 1 billion tokens (see~\cref{fig:trend}).

\begin{table*}[ht]
\begin{center}
\begin{tabular}{lc}
\toprule
 \textbf{Method} & \textbf{Computation Complexity}  \\
\midrule
Recurrent & $\mathcal{O}(Nd^2)$ \\
Vanilla Attention & $\mathcal{O}(N^2d)$ \\
Sparse Attention & $\mathcal{O}(N\sqrt{N}d)$ \\
\midrule
\bf Dilated Attention (This Work) & $\mathcal{O}(Nd)$ \\
\bottomrule
\end{tabular}
\caption{Comparison of computation complexity among different methods. $N$ is the sequence length and $d$ is the hidden dimension.}
\label{tab:complexity}
\end{center}
\end{table*}

\textbf{In this work, we successfully scale the sequence length to 1 billion tokens}. Our solution is \ours{}, which replaces the attention of vanilla Transformers with a novel component named dilated attention. The general design principle is - \emph{attention allocation decreases exponentially as the distance between tokens grows}. We prove that it obtains a linear computation complexity and a logarithm dependency between tokens. This deals with the contradiction between limited attention resources and the accessibility to every token. In the implementation, \our{} can be transformed into a dense Transformer, which seamlessly supports the off-the-shelf optimization for Transformers (e.g., kernel fusion, quantization, and distributed training).
Taking advantage of the linear complexity, \our{} can parallelize the training across nodes, breaking the constraint of both computation and memory with a distributed algorithm. This allows us to efficiently scale up the sequence length to 1B tokens with nearly constant runtime (see \cref{fig:runtime}), while vanilla Transformer suffers from quadratic complexity.

\section{\ours{}}

\subsection{Preliminary}

The core of Transformers~\cite{transformer} is self-attention, which maps a query and a set of keys and values to output. Given the inputs $Q,K,V \in \mathbb{R}^{N \times d}$, it computes the outputs $O$ with

\begin{equation}
    O = \text{softmax}(QK^{T})V
\end{equation}

Self-attention struggles with long sequences, due to its quadratic dependency on the sequence length. One query would attend to all keys and values, leading to computational inefficiencies.

Sparse attention alleviates this issue by restricting the query's access to a subset of keys and values. The key of sparse attention is the sparse attention pattern $S \in \{0, 1\}^{N \times N}$, which determines specific keys and values that the query $Q$ can attend to.

\begin{equation}
    O = \text{softmax}(QK^{T} \odot \mathbbm{1}_{S})V
\end{equation}

For example, the fixed pattern of sparse Transformer~\cite{sparsetransformer} is composed of a local pattern and a strided pattern. The sequence is divided into blocks of length $l$. The local pattern allows one query to attend to tokens within the same block, while strided pattern allows one query to attend to the last $c$ tokens of each block. Formally, the local pattern $S_i^{(1)} = \{ j \mid \lfloor j/l \rfloor = \lfloor i/l \rfloor \}$, and the strided pattern $S_i^{(2)} = \{ j \mid j \: \text{mod} \: l \in \{t, t+1,...,l\} \}$.

\subsection{Dilated Attention}
\label{sec:diattn}
\begin{figure}[t]
\centering

\includegraphics[width=1.0\columnwidth]{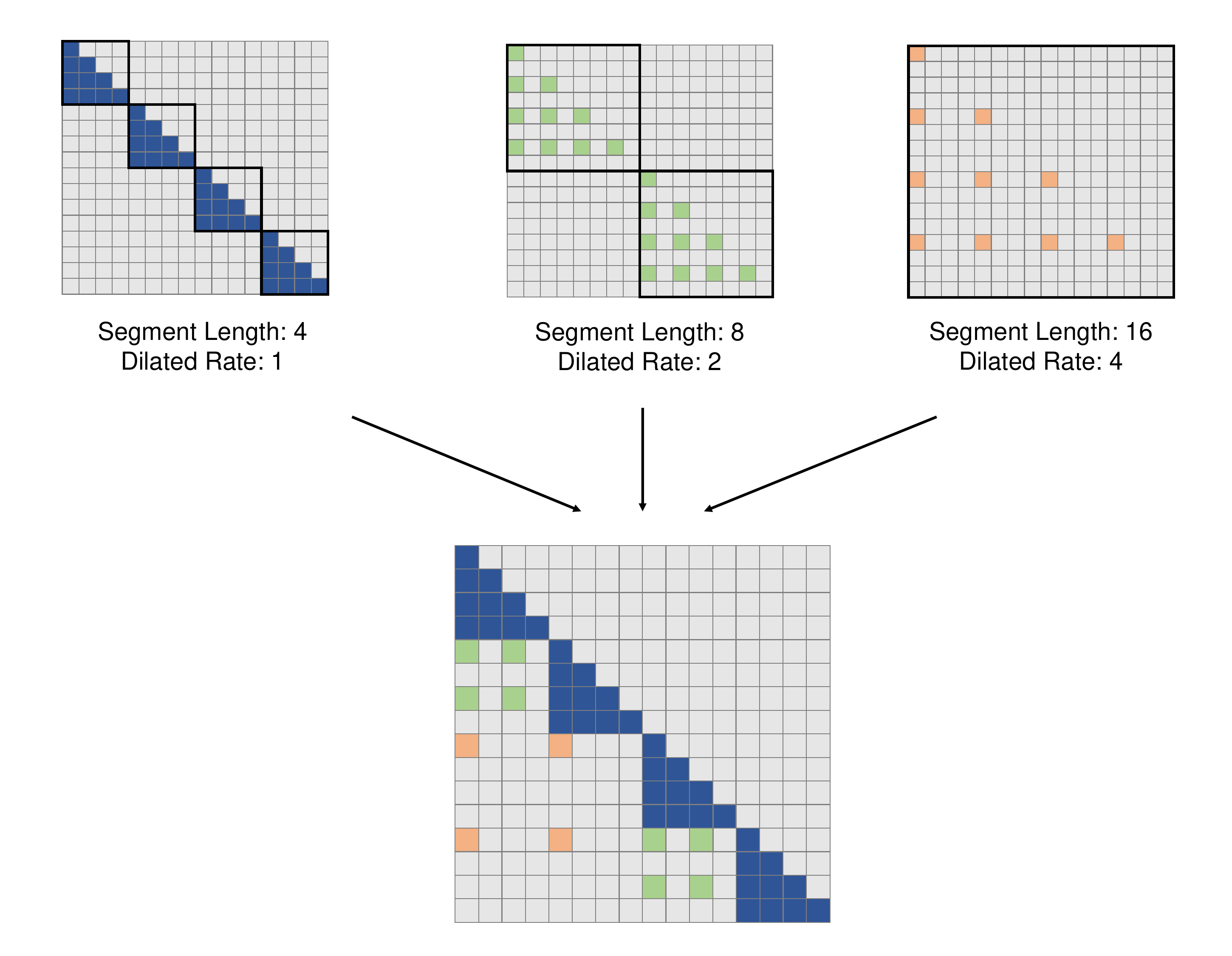}

\caption{Building blocks of dilated attention used in \our{}. It consists of a series of attention patterns for modeling both short-range and long-range dependency. The number of attention patterns can be extended according to the sequence length.
}
\label{fig:dit}
\end{figure}

\cref{fig:dit} illustrates the overview of dilated attention.
Dilated attention splits the input ($Q$, $K$, $V$) into segments $\{(\widetilde{Q}_{i},\widetilde{K}_{i},\widetilde{V}_{i})\}^{\frac{N}{w}}$ equally with a segment length $w$. Each segment is then sparsified along the sequence dimension by selecting the rows with an interval $r$. The computation can be written as:

\begin{equation}
    \widetilde{Q}_{i} = [Q_{iw}, Q_{iw+r}, Q_{iw+2r}, ..., Q_{(i+1)w-1}]
    \label{eq:3}
\end{equation}

\begin{equation}
    \widetilde{K}_{i} = [K_{iw}, K_{iw+r}, K_{iw+2r}, ..., K_{(i+1)w-1}]
\end{equation}

\begin{equation}
    \widetilde{V}_{i} = [V_{iw}, V_{iw+r}, V_{iw+2r}, ..., V_{(i+1)w-1}]
    \label{eq:5}
\end{equation}

The sparsified segments $\{(\widetilde{Q}_{i},\widetilde{K}_{i},\widetilde{V}_{i})\}^{\frac{N}{w}}$ are fed into the attention in parallel, after which are scattered and concatenated as the output $O$:

\begin{equation}
    \widetilde{O}_{i} = \text{softmax}(\widetilde{Q}_{i} \widetilde{K}_{i}^{T})\widetilde{V}_{i}
\end{equation}

\begin{equation}
    \hat{O}_i = \{\widetilde{O}_{i,j} | j \: \text{mod} \: r=0; 0 | j \: \text{mod} \: r \neq 0\}
\end{equation}

\begin{equation}
    O = [\hat{O}_{0}, \hat{O}_{1}, ..., \hat{O}_{\frac{N}{w}-1}]
    \label{eq:8}
\end{equation}

In the implementation, the dilated attention can be transformed into dense attention between a gathering operation over the input $(Q, K, V)$ and a scattering operation over the output $\widetilde{O}_{i}$, so it can directly reuse any optimization for vanilla attention (e.g., flash attention~\cite{flashattn}). Dilated attention can significantly reduce the computation cost by a factor of $\frac{N}{w}r^2$ over the vanilla attention.

In practice, the segment size $w$ trades the globality of attention for efficiency, while the dilation with a size $r$ reduces the computation cost by approximating the attention matrix. To capture both long-range and short-range information efficiently, we implement a mixture of dilated attentions with different segment sizes and dilation rates $\{r_i, w_i\}^k$:

\begin{equation}
    O = \sum_{i=1}^{k} \alpha_i O|_{r_i,w_i}
\end{equation}

\begin{equation}
    \alpha_i = \frac{s_i}{\sum_j s_j}
\end{equation}

where $s_i$ denotes the denominator of the attention softmax for $O|_{r_i,w_i}$. Note that the computations for $\{O|_{r_i,w_i}\}^k$ are in parallel because there is no computation dependency among them. Experiments show that dynamic weights calculated by the denominator of the attention softmax are better than learnable fixed weights. For a query attends to keys in different dilated attentions, our method to mix dilated attentions is equivalent to gather keys in different parts and calculate softmax together.

Intuitively, the local attention should be precisely computed, while the global attention can be approximate. Therefore, we set a larger $w_i$ with a bigger $r_i$. Moreover, we gradually increase the $w_i$ for each attention until it reaches the maximum length $N$ or the number of attention patterns $k$:

\begin{equation}
    w=\{w_0, w_1, w_2, ..., N\}^k \quad (w_i<w_{i+1}<N)
    \label{eq:w}
\end{equation}

\begin{equation}
    r=\{1, r_1, r_2, ..., r_k\}^k \quad (1<r_i<r_{i+1})
    \label{eq:r}
\end{equation}

In practice, we set $w$ and $r$ to geometric sequences for an exponential attentive field.

\subsection{Multi-Head Dilated Attention}

\begin{figure}[t]
\centering

\includegraphics[width=1.0\columnwidth]{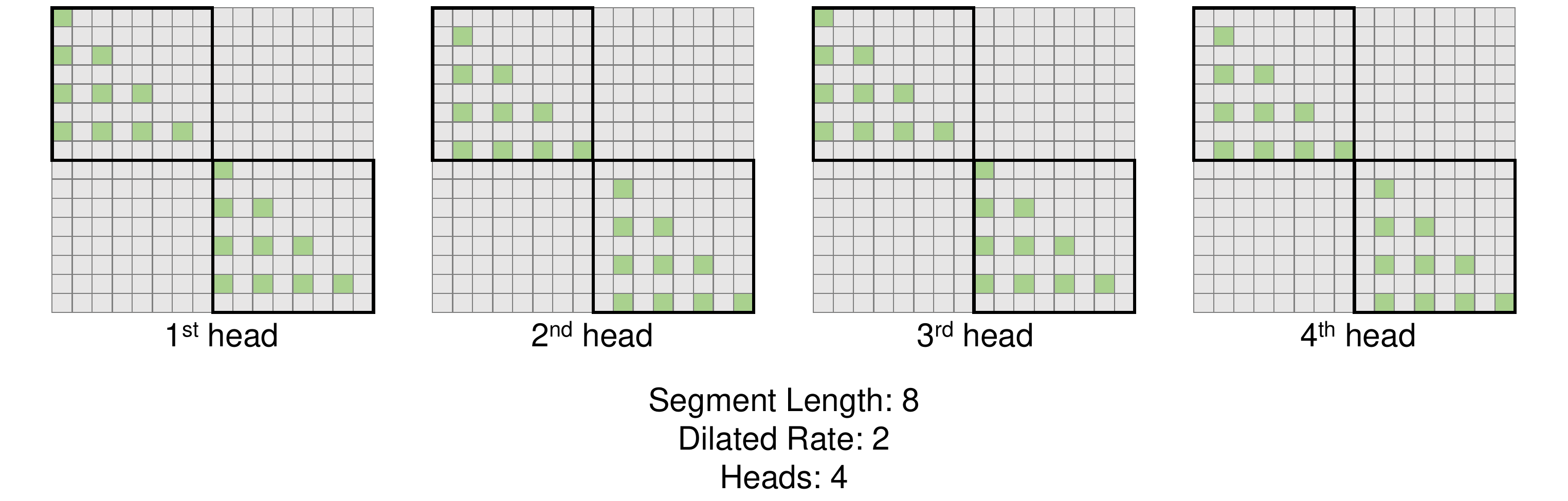}

\caption{Dilated attention with multiple heads. The attention patterns differ among heads by shifting the position successively.
}
\label{fig:multihead}
\end{figure}

As shown in~\cref{fig:multihead}, we differ in the computation among different heads by sparsifying different parts of the query-key-value pairs.
Specifically, for the $j$-th head, we have an offset $s_j = j \: \text{mod} \: r $ when selecting the $(Q, K, V)$:

\begin{equation}
    \widetilde{Q}_{i} = [Q_{iw+s_j}, Q_{iw+s_j+r}, Q_{iw+s_j+2r}, ..., Q_{(i+1)w+s_j-1}]
\end{equation}

\begin{equation}
    \widetilde{K}_{i} = [K_{iw+s_j}, K_{iw+s_j+r}, K_{iw+s_j+2r}, ..., K_{(i+1)w+s_j-1}]
\end{equation}

\begin{equation}
    \widetilde{V}_{i} = [V_{iw+s_j}, V_{iw+s_j+r}, V_{iw+s_j+2r}, ..., V_{(i+1)w+s_j-1}]
\end{equation}

Following the vanilla multi-head attention, the outputs of different heads are concatenated into a final output. The rest of the computation remains the same as the single-head counterpart in~\cref{sec:diattn}.

\subsection{Computational Complexity and Token Dependency}

Given dilated attention with a segment size and dilation rate of $(r, w)$, each query-key-value pair is sparsified from $(Q,K,V) \in \mathbb{R}^{N \times d}$ to $(Q,K,V) \in \mathbb{R}^{\frac{w}{r} \times d}$, so the flops of the attention computation are estimated as:
\begin{equation}
    FLOPs=\frac{2N}{w}(\frac{w}{r})^2d=\frac{2Nwd}{r^2}
\end{equation}

We further extend it to dilated attention with multiple segment sizes and dilation rates. The flops can be written as:
\begin{equation}
    FLOPs=2Nd \sum_{i=1}^{k}{\frac{w_i}{r_i^2}}
\end{equation}

With the segment sizes and dilation rates in~\cref{eq:w} and~\cref{eq:r}, the flops are given by
\begin{equation}
    FLOPs=2w_{0}Nd \sum_{i=0}^{k-1}{\frac{1}{\alpha^i}} \leq \frac{2\alpha}{\alpha - 1} w_{0}Nd \quad (\alpha > 1)
\end{equation}
where $w_0$ is a predefined constant and $\alpha$ is the common ratio for geometric sequences $w$ and $r$. Therefore, the computation complexity of dilated attention is approximate to $\mathcal{O}(Nd)$.

Moreover, the information of each tokens can be propagated to a maximum distance of $D$:
\begin{equation}
    D=\sum_{i=0}^{l-1}w_i=w_0\sum_{i=0}^{l-1}\alpha^i \approx \frac{w_0}{\alpha-1}\alpha^l
\end{equation}
where $l$ is the length of the propagated path. Therefore, the maximum path length of a sequence with $N$ tokens can be estimated as:
\begin{equation}
    L \approx \log_{\alpha} \frac{N(\alpha-1)}{w_0} \quad (\alpha > 1)
\end{equation}
This proves that the token dependency is approximate to $\mathcal{O}(\log N)$.

\section{\ours{} as a Distributed Trainer: Scaling up to 1B Tokens}

Although the computation complexity of dilated attention has been greatly reduced to $\mathcal{O}(Nd)$, it is infeasible to scale the sequence length to the million level on a single GPU device due to the computation and memory constraints. There are some distributed training algorithms for large-scale model training, such as model parallelism~\cite{megatron}, sequence parallelism~\cite{sequenceparallel,reduceact}, and pipeline parallelism~\cite{gpipe}. However, they are insufficient for \our{} especially when the sequence dimension is extremely large.

\subsection{Distributed Algorithm}

\begin{figure}[t]
\centering

\includegraphics[width=0.6\columnwidth]{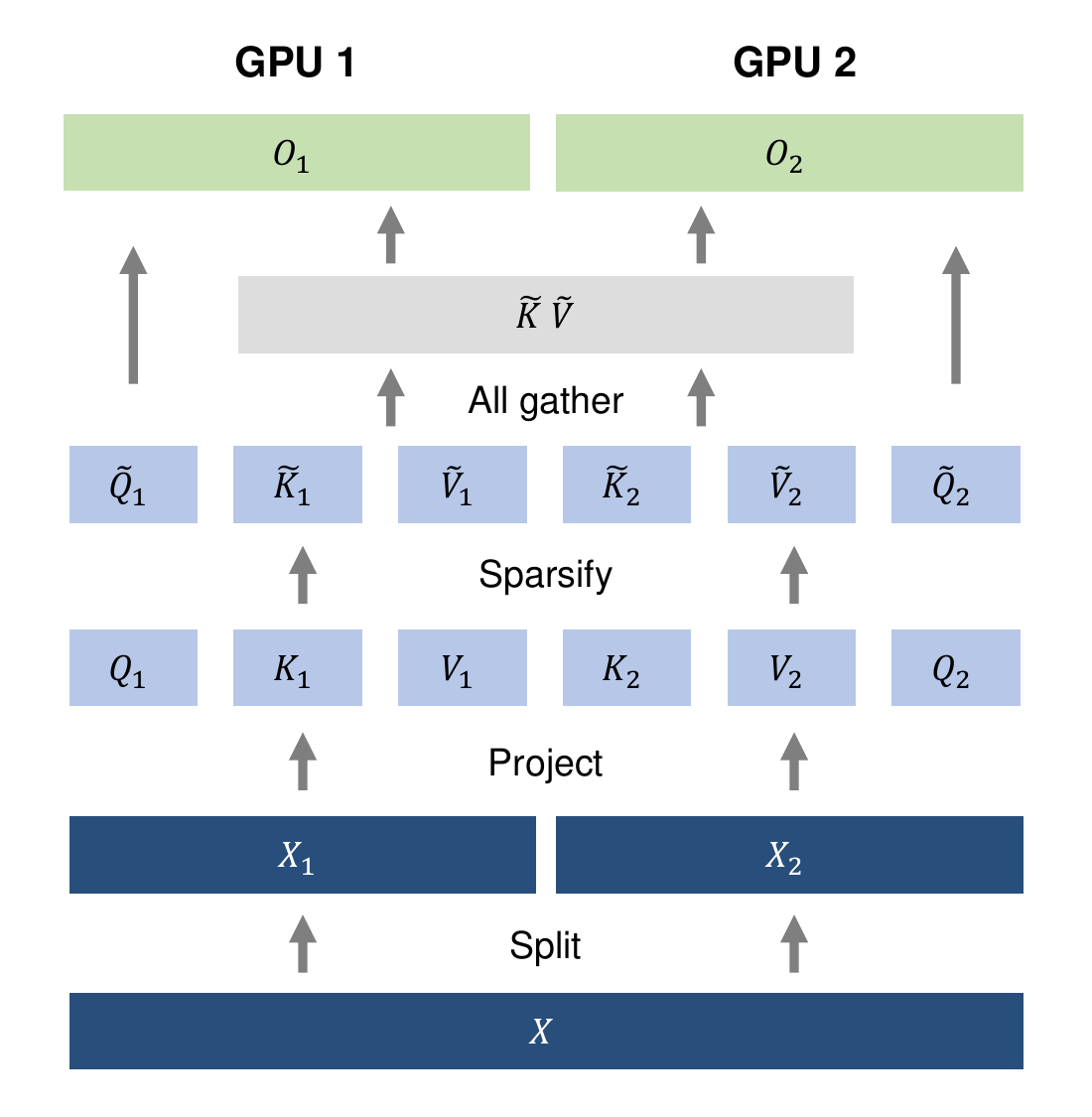}

\caption{Distributed training of \our{} on two GPU devices. It parallelizes the training by partitioning the sequence dimension. The computation and communication costs are nearly constant as the number of devices grows.
}
\label{fig:distributed}
\end{figure}

We take advantage of the linear computation complexity of \our{} for the distributed training of sequence dimension.
Without loss of generality, \cref{fig:distributed} presents our distributed algorithm on two GPUs, which can be further scaled to an arbitrary number of devices.
We start by splitting the input sequence along the sequence dimension. Each sequence is put on one device separately:

\begin{equation}
    X = [X_{1}, X_{2}]
\end{equation}

Then, they are projected into queries, keys, and values on the two devices:

\begin{equation}
    [Q_1, K_1, V_1] = [W_Q, W_K, W_V] X_1, \quad [Q_2, K_2, V_2] = [W_Q, W_K, W_V] X_2
\end{equation}

For the segment length $w_i \leq l$ (where $l$ is the sequence length on the local device), we compute the attention locally with~\cref{eq:3} to~\cref{eq:8}. For the segment length $w_i > l$, the keys and values are distributed across different devices. Therefore, we collect the key-value pairs before computing the attention. We use~\cref{eq:3} to~\cref{eq:5} to sparsify the $\{Q, K, V\}$ into $\{\widetilde{Q}, \widetilde{K}, \widetilde{V}\}$. An all-gather operation is implemented to collect the key-value pairs:

\begin{equation}
    \widetilde{K} = [\widetilde{K_1}, \widetilde{K_2}], \quad \widetilde{V} = [\widetilde{V_1}, \widetilde{V_2}]
\end{equation}

Note that the all-gather operation in the backward becomes a reduce-scatter operation. Different from vanilla attention, both sizes of $\widetilde{K_i}$ and $\widetilde{V_i}$ are independent of the sequence length $N$, making the communication cost constant.

Finally, we compute the cross-attention with the local queries $\widetilde{Q_i}$ and the global key-value pairs $\{\widetilde{K}, \widetilde{V}\}$. The formulation is written as:

\begin{equation}
    \widetilde{O_1} = \text{softmax}(\widetilde{Q_1} \widetilde{K}^{T})\widetilde{V}, \quad
    \widetilde{O_2} = \text{softmax}(\widetilde{Q_2} \widetilde{K}^{T})\widetilde{V}
\end{equation}

The concatenation of the outputs across different devices becomes the final attention output:

\begin{equation}
    \widetilde{O} = [\widetilde{O_1}, \widetilde{O_2}]
\end{equation}

The distributed algorithm described above is orthogonal to other parallelisms, including data parallelism which partitions the batch dimension, model parallelism which partitions the hidden dimension, and pipeline parallelism which partitions the layers.

\subsection{Scaling up to 1B Tokens}

\begin{figure}[t]
\centering
\includegraphics[width=0.85\columnwidth]{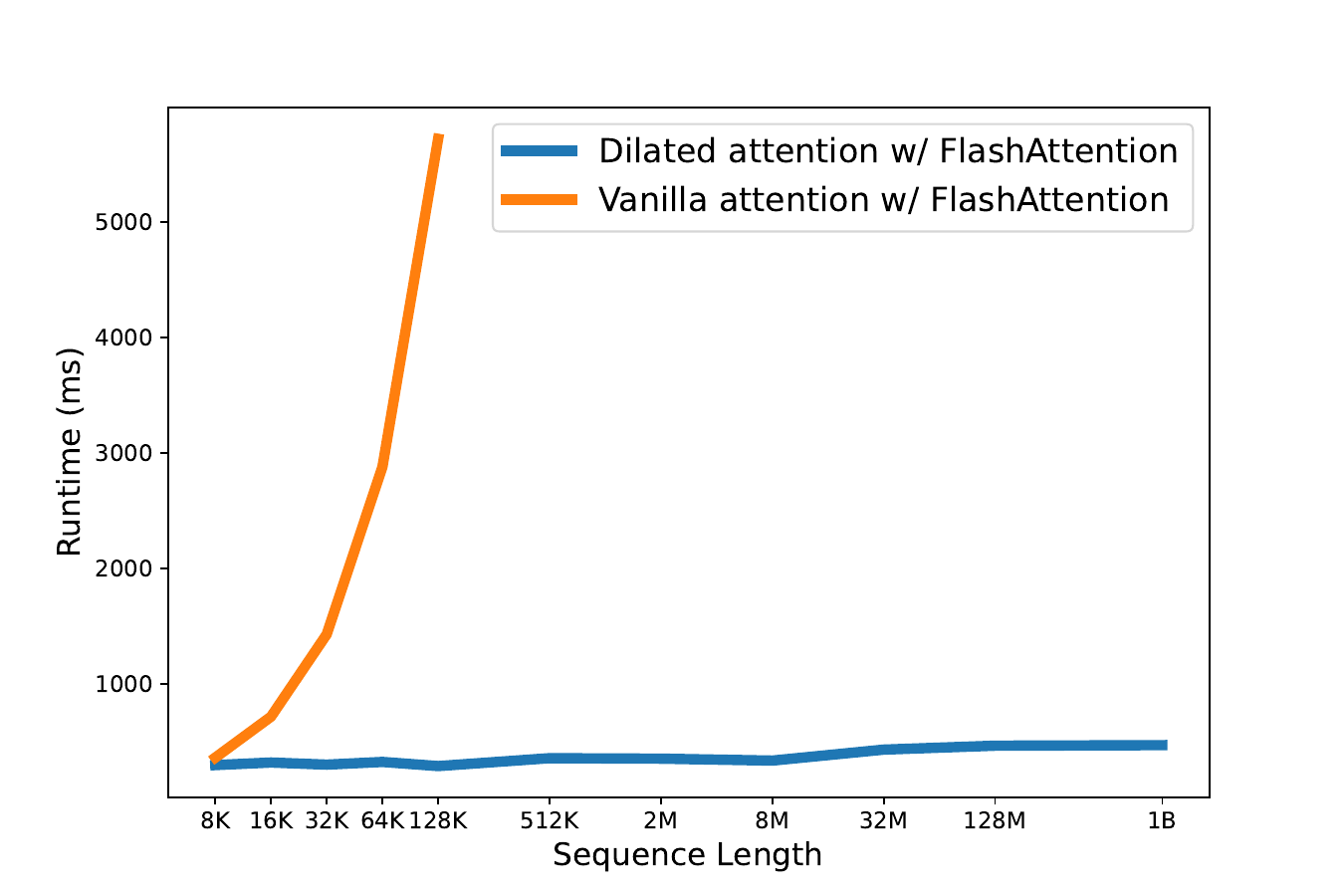}
\caption{Runtime of our dilated attention and vanilla attention. Both are equipped with FlashAttention~\cite{flashattn}.
}
\label{fig:runtime}
\end{figure}

We verify the feasibility of scaling to 1B tokens with the modern distributed systems.
Starting from 8K, we gradually scale the sequence length until the limit of GPU memory.
We reduce the batch size accordingly to keep the number of tokens per batch at 1 billion.
Each model of different sequence lengths has up to 3 segment lengths, which are 2,048, the number of tokens per device, and the sequence length.
We compute the average speed in the forward propagation for 10 different runs.

\cref{fig:runtime} reports the runtime of vanilla attention and our dilated attention. Both of them are implemented with FlashAttention Kernel for saving memory and improving speed. It shows that dilated attention can successfully scale up the sequence length with almost constant latency. By partitioning the sequence dimension, it can leverage the distributed systems to scale the sequence length to 1 billion tokens. In contrast, vanilla attention suffers from the quadratic dependency on the sequence length. Its latency dramatically increases as the length grows. Moreover, there is no distributed algorithm for vanilla attention to break sequence length limitation. This proves the advantage of the linear complexity as well as the distributed algorithm for \our{}.

\section{Experiments on Language Modeling}

\subsection{Setup}

We implement \ours{} on language modeling. The backbone architecture is \textsc{Magneto}~\cite{magneto} with \textsc{xPos}~\cite{xpos} relative position encoding, except that we replace the standard attention with our dilated attention. We use the base-size configuration of \textsc{Magneto}, which has a hidden dimension of 768, 12 attention heads, and 12 decoder layers. We pre-train the model with The Stack dataset~\cite{stack}, a source code collection in over 300 programming languages. The data is preprocessed with the tiktoken tokenizer\footnote{\url{https://github.com/openai/tiktoken}} with \texttt{cl100k\_base} encoding.
The models are trained with a batch size of 0.5M tokens for 300K steps.
More details regarding the hyperparameters can be found in the appendix. All experiments are conducted based on the \emph{torchscale}~\citep{torchscale} codebase.

\subsection{Results}

We compare \ours{} with both vanilla Transformer and sparse Transformers. The differences among the architectures are the attention layers, while the others remain the same. We scale the sequence length of these models from 2K to 32K, while reducing the batch size to keep the number of tokens per batch constant. For \ours{}, we use segment lengths of $w=\{2048, 4096, 8192, 16384, 32768\}$, and the dilated ratios are $r=\{1,2,4,6,12\}$. We implement the fixed pattern for sparse attention as in~\cite{sparsetransformer} with multiple heads attending to distinct subblocks. The block size is set to 2048. We adjust their sparse ratios to match the computation flops with \our{} so that the comparison is fair. The attention layers in vanilla Transformers are dense and fully connected, so the computation cost is much higher. Due to the computation constraints, we only scale it up to 32K sequence length. All of our implementations of attention variants are based on FlashAttention\footnote{\url{https://github.com/HazyResearch/flash-attention/tree/main}} for training efficiency. We customize the flash attention kernels for both sparse attention and dilated attention.

\begin{table*}[t]
\setlength{\tabcolsep}{11pt}
\centering
\begin{tabular}{lcc|ccc}
\toprule
\multirow{2}{*}{\textbf{Model}} & \multirow{2}{*}{\textbf{Length}} & \multirow{2}{*}{\textbf{Batch}}  & \multicolumn{3}{c}{\textbf{Github}}  \\
& & & \textbf{2K} & \textbf{8K} & \textbf{32K} \\
\midrule
Transformer~\cite{transformer} & 2K & 256 &  4.24 & 5.07 & 11.29\\
\midrule
Sparse Transformer~\cite{sparsetransformer} & \multirow{2}{*}{8K} & \multirow{2}{*}{64} &  4.39 & 3.35 & 8.79 \\
\bf \our{} (ours) &   &  & 4.23 & 3.24 & 3.36 \\
\midrule
Sparse Transformer~\cite{sparsetransformer} & \multirow{2}{*}{16K} & \multirow{2}{*}{32}  &  4.85 & 3.73 &  19.77 \\
\bf \our{} (ours) & &   & 4.27 & 3.26 & 3.31 \\
\midrule
Sparse Transformer~\cite{sparsetransformer} & \multirow{2}{*}{32K} & \multirow{2}{*}{16}   & 5.15 & 4.00 & 3.64 \\
\bf \our{} (ours) &  &  & 4.37 & 3.33 & 3.01 \\
\bottomrule
\end{tabular}
\caption{Perplexity of language models for \our{} and the baselines.}
\label{tab:results}
\end{table*}

\cref{tab:results} summarizes the results of these models on the Stack dataset. We use perplexity as the evaluation metric. 
The models are tested with different sequence lengths, ranging from 2K to 32K. When the input is longer than the maximum length that the models support, we implement blockwise causal attention (BCA)~\cite{xpos}, a state-of-the-art extrapolation method for language model inference. Besides, we remove the absolute position encoding. Primarily, the results demonstrate that increasing the sequence length during training generally leads to a better language model. Secondly, the extrapolation of sequence length in inference does not apply to the case when the length is much larger than the model supports. Finally, \our{} consistently outperforms the baseline models, proving its effectiveness in language modeling.

\begin{figure}[t]
\centering

\includegraphics[width=0.8\columnwidth]{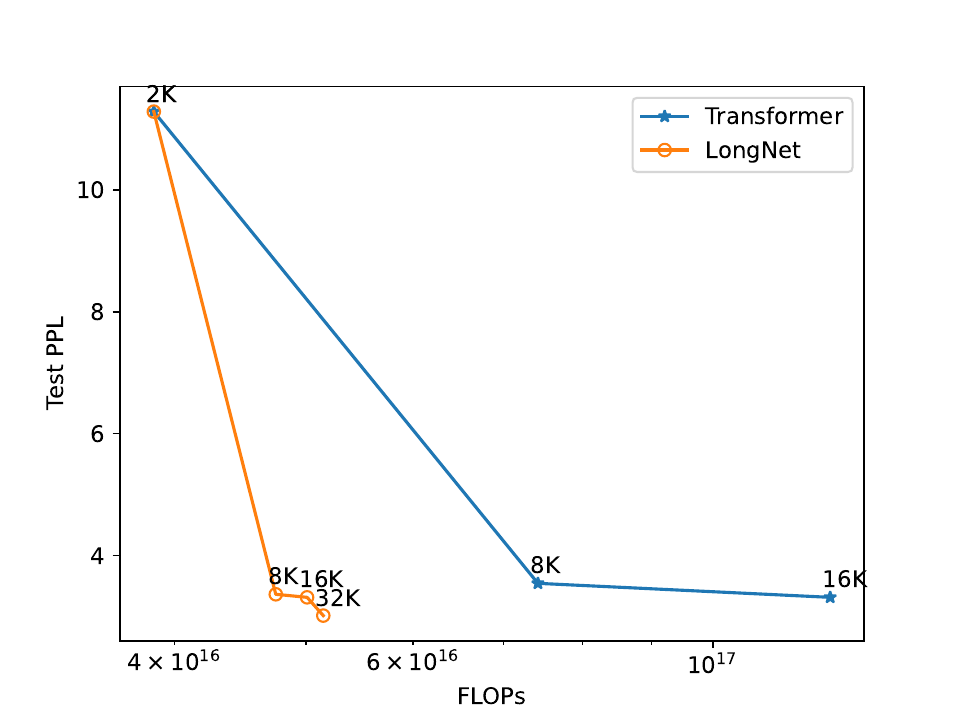}

\caption{Test perplexity of \our{} and dense Transformers using different sequence lengths during training. \our{} outperforms dense Transformers with a lower perplexity and a significantly smaller amount of computation.
}
\label{fig:scaling_length}
\end{figure}

\subsection{Scaling Curves of Sequence Length}

Previous work~\cite{scalinglaw} has shown that language models follow some scaling laws by increasing parameters or training tokens. We are interested in the performance of language models when the context length is scaled up during training. We test the losses with inputs of a mixture of different lengths, from 1K to 32K. We use blockwise causal attention during inference to improve the generalization of sequence lengths.

\cref{fig:scaling_length} plots the scaling curves of sequence length for both vanilla Transformers and \our{}. We estimate the amount of compute by calculating the total flops of matrix multiplication. 
The results show that both vanilla Transformers and \our{} benefit from a larger context length during training. However, \our{} can scale up the context length more efficiently, achieving a lower test loss with a smaller amount of computing. This demonstrates the advantage of longer training input over extrapolation.
In conclusion, our experiments show that \our{} is a more efficient way to scale up the context length in language models. This is because \our{} can learn longer-range dependencies more effectively.

\subsection{Scaling up Model Size}

\begin{figure}[t]
\centering
\subfigure[]{
\includegraphics[width=0.46\columnwidth]{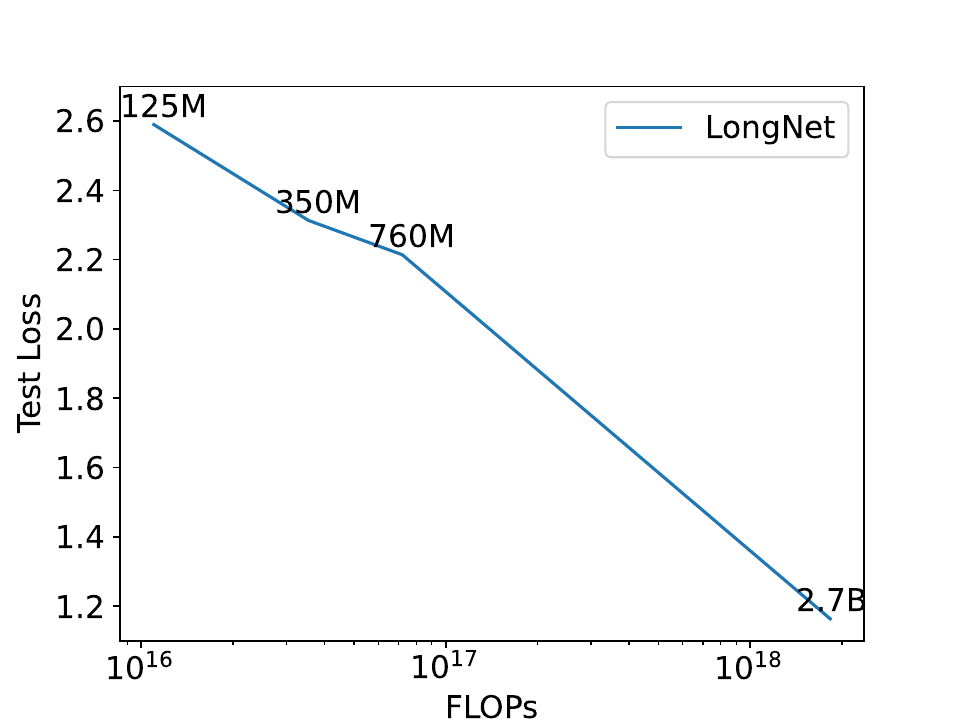}
\label{fig:scaling_law}
}
\subfigure[]{
\includegraphics[width=0.46\columnwidth]{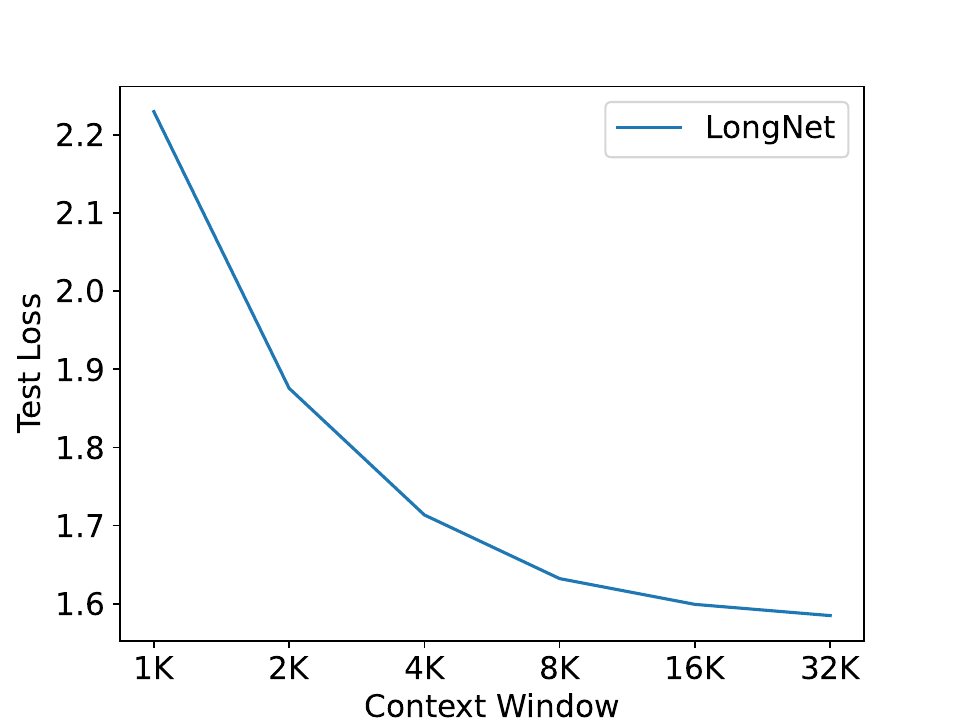}
\label{fig:long_prompt}
}
\caption{\textbf{Left:} Test loss of \our{} with an increasing model size. The scaling curve follows a similar law to the vanilla Transformers. \textbf{Right:} Test loss of \our{} using different context windows. A longer context window yields better language modeling.
}
\end{figure}

An important property of large language models is that the loss scales as a power law with compute. To verify whether \our{} still follows the similar scaling law, we train a series of models with different model sizes, from 125 million to 2.7 billion parameters. The 2.7B model is trained with 300B tokens, while the rest digest about 40B tokens. \cref{fig:scaling_law} plots the scaling curve of \our{} regarding the compute. We compute the perplexity on the same test set. The amount of compute is estimated by calculating the total flops of matrix multiplication during training. It proves that \our{} can still follow the power law. This implies that the dense Transformer is not a prerequisite for scaling the language models. Additionally, the scalability and the efficiency are both obtained by \our{}. 
 
\subsection{Long Context Prompting}

Prompting is an essential method to guide and provide additional information to the language models. We conduct experiments to verify whether \ours{} can benefit from a longer context window for prompting. Specifically, we reserve a piece of prefixes as the prompt and test the perplexity of its suffixes. We gradually scale the length of the prompt from 2K to 32K.
For a fair comparison, we keep the suffixes the same, while increasing the length of the prefixes to the maximum lengths of the models. The results on the test set are reported in \cref{fig:long_prompt}. It shows that the test loss of \our{} gradually decreases as the context window grows. This demonstrates the superiority of \our{} in fully leveraging the long context to improve the language model.

\section{Conclusion and Future Work}

We present \ours{}, a Transformer variant that can scale the sequence length to 1 billion tokens and beyond, with no loss in shorter sequences. The core of \ours{} is dilated attention, which reduces the computation complexity from quadratic to linear. \our{} can be served as a distributed trainer that parallelizes the training of a sequence across multiple GPU devices. Experiments show that \our{} has superior performance over the strong baselines on modeling both long and short sequences. In the future, we will extend \our{} to support more tasks, e.g., multimodal large language modeling~\citep{kosmos-1,kosmos-2}, BEiT pretraining~\citep{beit,beitv2,beit3}, and genomic data modeling.

\paragraph{Acknowledgement}
We would like to acknowledge Yuqing Xia and Jilong Xue for the early exploration of the flash attention kernel.

\nocite{logsparse}

\bibliographystyle{alpha}
\bibliography{longnet}

\newpage

\appendix

\section{Hyperparameters}

\begin{table}[ht]
\centering
\begin{tabular}{lc}
\toprule
\bf Hyperparameters & \bf Value \\
\midrule
Layers & 12 \\
Hidden size & 768 \\
FFN size & 3072 \\
Heads & 12 \\
\midrule
Learning rate & 6e-4 \\
LR scheduler & Polynomial decay \\
Warm-up steps & 750 \\
Tokens per batch & 500K \\
Adam $\beta$ & (0.9, 0.98) \\
Training steps & 300K \\
\midrule
Gradient clipping & 2.0 \\
Dropout & 0.0 \\
Weight decay & 0.01 \\
\bottomrule
\\
\end{tabular}
\caption{Hyperparamters used for the models in~\cref{tab:results}.
}
\end{table}

\begin{table}[ht]
\centering
\begin{tabular}{ccccccccc}
\toprule
\bf Parameters & \bf Layers & \bf Hidden & \bf Heads & \bf Learning Rate & \bf Batch Size \\
\midrule
125M & 12 & 768 & 12 & 6e-4 & 500K \\
350M & 24 & 1024 & 16 & 6e-4 & 500K \\
760M & 24 & 1536 & 16 & 6e-4 & 500K \\
2.7B & 32 & 2560 & 32 & 2e-4 & 4M \\ 
\bottomrule
\\
\end{tabular}
\caption{Hyperparamters used for the experiments in~\cref{fig:scaling_law}.
}
\end{table}

\end{document}